\pdfoutput=1
\documentclass[11pt]{article}

\usepackage[final]{acl}

\usepackage{times}
\usepackage{latexsym}

\usepackage{booktabs}
\usepackage{multirow} 
\usepackage[T1]{fontenc}
\usepackage[utf8]{inputenc}

\usepackage{microtype}

\usepackage{inconsolata}

\usepackage{graphicx}

\usepackage[most]{tcolorbox}
\usepackage{booktabs}
\usepackage{geometry}
\usepackage{enumitem}
\title{Table-Text Alignment: Explaining Claim Verification Against Tables in Scientific Papers
}

\author{
Xanh Ho,$^1$ 
Sunisth Kumar,$^2$ 
Yun-Ang Wu,$^{3*}$ \\
\textbf{Florian Boudin},$^4$ 
\textbf{Atsuhiro Takasu},$^1$ \and
\textbf{Akiko Aizawa}$^{1,2}$ \\ 
$^1$National Institute of Informatics, Japan \hspace{1cm} $^2$The University of Tokyo, Japan \\
$^3$National Taiwan University \hspace{1cm}
$^4$JFLI, CNRS, Nantes Université, France \\ 
{\tt \{xanh, takasu, aizawa\}@nii.ac.jp} \hspace{1cm}
{\tt sunisth@g.ecc.u-tokyo.ac.jp} \\
{\tt r11944072@csie.ntu.edu.tw} \hspace{1cm}
{\tt florian.boudin@univ-nantes.fr} 
}

\begin{document}
\maketitle

\begingroup\def\thefootnote{*}\footnotetext{Research conducted during internship at NII, Japan.}\endgroup

\begin{abstract}
% Abstract Florian's verion:

Scientific claim verification against tables typically requires predicting whether a claim is supported or refuted given a table.
However, we argue that predicting the final label alone is insufficient: it reveals little about the model’s reasoning and offers limited interpretability. 
To address this, we reframe table–text alignment as an explanation task, requiring models to identify the table cells essential for claim verification.
We build a new dataset by extending the SciTab benchmark with human-annotated cell-level rationales.
Annotators verify the claim label and highlight the minimal set of cells needed to support their decision. 
After the annotation process, we utilize the collected information and propose a taxonomy for handling ambiguous cases.
Our experiments show that (i) incorporating table alignment information improves claim verification performance, and (ii) most LLMs, while often predicting correct labels, fail to recover human-aligned rationales, suggesting that their predictions do not stem from faithful reasoning.\footnote{Our data and code are available at \url{https://github.com/Alab-NII/SciTabAlign}
% \footnote{
% Our prompts and dataset are included in the submission’s .zip file.
% We will release them publicly in the future.
}

\end{abstract}

\section{Introduction}

% Background 
%
Claim verification against tables requires models to determine whether a natural language claim is supported or refuted based on structured tabular data.
Several benchmarks have been proposed in the general domain, such as TabFact~\cite{Chen2020TabFact}, INFOTABS~\cite{gupta-etal-2020-infotabs}, and FEVEROUS~\cite{aly2021feverous}, primarily focusing on Wikipedia tables.
However, tables in scientific papers pose additional challenges: they are often denser, more structured, and require domain-specific reasoning.
%

% Background & Motivations
% SEM is a shared task that includes two subtasks: claim verification and cells evidence selection.
% The two main issues with SEM are that its claims are simple and generated through crowdsourcing, which means they do not accurately reflect the original claims found in scientific papers.
% In contrast, SciTab focuses solely on the claim verification task.
% %
Two datasets have recently addressed this task in the scientific domain: SEM-TAB-FACTS~\cite[SEM;][]{wang-etal-2021-semeval}, which includes both claim verification and cell-level evidence selection, and SciTab~\cite{lu-etal-2023-scitab}, which focuses solely on claim verification. 
While SEM includes an alignment component, its claims are crowd-generated and simplified, limiting their representativeness.
SciTab, in contrast, uses naturally occurring claims but lacks explicit annotations that explain why a given label is correct.

% However, we argue that predicting the final label alone, as SciTab does, is insufficient for evaluating a model's true understanding ability and also limits the explainability of its predictions.
We argue that label prediction alone, as in SciTab, is not enough.
It fails to reveal whether a model truly understands the table content, nor does it provide interpretable reasoning.
For both evaluation and practical use, models need to go beyond classification and provide explanations grounded in tabular evidence.

% Table-text alignment 
%On the other hand, from the perspective of assisting researchers in reading scientific papers~\cite{lo2023semanticreaderprojectaugmenting}, table–text alignment can help identify which parts of a table are referenced in a given sentence.
%This alignment allows researchers to more quickly and accurately locate relevant table content, enhancing reading efficiency and comprehension.
%If such alignment data can be obtained, it has the potential to support and streamline the human reading process.

From the perspective of scientific reading tools~\cite{lo2023semanticreaderprojectaugmenting}, table–text alignment is also crucial.
It allows readers to quickly locate which parts of a table are referenced in the text, improving comprehension and accelerating the reading process. 
Such alignments could directly support scientific workflows by making tabular evidence more accessible and interpretable.

% What we do
%To address these issues, we propose formulating table–text alignment as an explanation task for scientific claim verification.
%Specifically, we enhance the existing SciTab dataset by adding an explanation task in the form of table–text alignment, which requires models to identify the specific cells, rows, or columns in the table that are necessary for predicting the claim label.
To address these limitations, we reframe table–text alignment as an explanation task for scientific claim verification.
Specifically, we extend the SciTab dataset with human-annotated cell-level rationales.
For each claim–table pair, annotators verify the claim label and highlight the minimal set of table cells needed to support the decision.
%
%We manually construct our dataset by having annotators verify claim labels and highlight the table cells essential for determining those labels. 
%During annotation, we propose a comprehensive taxonomy categorizing ambiguous cases in scientific claim verification against tables. 
%Our taxonomy identifies five types of ambiguity: (i) Table Conversion Errors, (ii) Additional Context Requirements, (iii) Unexpected Claim Types, (iv) Subjective Adjectives, and (v) Unclear Claims. 

During annotation, we frequently encountered ambiguous cases in claim interpretation and evidence selection. 
To capture these edge cases systematically, we introduce a taxonomy of five ambiguity types in scientific table-based verification: (i) Table Conversion Errors, (ii) Additional Context Requirements, (iii) Unexpected Claim Types, (iv) Subjective Adjectives, and (v) Unclear Claims.
%
%We hope this taxonomy will guide future dataset construction by highlighting critical cases that need careful attention.
%
%We hope this taxonomy will serve as a reference for future dataset design and evaluation, encouraging more precise definitions and coverage of challenging instances.

% Results & Findings 

We use our dataset to evaluate various types of large language models (LLMs), including table-based models, open-source LLMs, and closed-source LLMs.
Our experiments also incorporate three different prompting strategies.
On average, our human-annotated cell-level rationales help improve the performance of the claim label prediction task.
The results show that while models achieve high macro-F1 scores on the claim label prediction task, their performance on the cell selection task remains low—even for advanced models like GPT-4o.
The highest score, 50.8, is achieved by Qwen 2.5 72B using CoT prompting.
Further analysis of the correlation between the two tasks reveals that although LLMs often correctly predict the claim label, their ability to identify the corresponding explanation cells is still limited.

\section{Related Work}

% Exsiting datasets 
%Claim verification, or fact-checking, has become increasingly important as it spans multiple domains, including news (e.g., LIAR~\cite{wang-2017-liar}), Wikipedia (e.g., FEVER~\cite{thorne-etal-2018-fever} and HoVer~\cite{jiang-etal-2020-hover}), scientific papers (e.g., SciFact~\cite{wadden-etal-2020-fact} and CLAIMCHECK~\cite{ou2025claimcheckgroundedllmcritiques}), and the medical domain (e.g., PubHealth~\cite{kotonya-toni-2020-explainable} and HealthFC~\cite{vladika-etal-2024-healthfc}).
%
Claim verification has been studied across multiple domains, including news~\cite{wang-2017-liar}, Wikipedia~\cite{thorne-etal-2018-fever,jiang-etal-2020-hover}, scientific literature~\cite{wadden-etal-2020-fact,ou2025claimcheckgroundedllmcritiques}, and medicine~\cite{kotonya-toni-2020-explainable,vladika-etal-2024-healthfc}.
%
%Additionally, several datasets have been created for claim verification involving structured or multimodal data sources, such as tables (e.g., TabFact~\cite{Chen2020TabFact} and SciTab~\cite{lu-etal-2023-scitab}), 
%figures (e.g., ChartCheck~\cite{akhtar-etal-2024-chartcheck}), 
%knowledge graphs (e.g., FactKG~\cite{kim-etal-2023-factkg}), and multimodal data (e.g., MMSci~\cite{yang2025doestablesourcematter}).
%
Beyond plain text, recent efforts have extended claim verification to structured or multimodal evidence, including tables~\cite{Chen2020TabFact,lu-etal-2023-scitab}, figures~\cite{akhtar-etal-2024-chartcheck}, knowledge graphs~\cite{kim-etal-2023-factkg} and multimodal data~\cite{yang2025doestablesourcematter}.

% Works do the same as we do 
%SEM-TAB-FACTS \cite[SEM;][]{wang-etal-2021-semeval} and TabEvidence \cite{gupta-etal-2022-right} are the datasets most relevant to our work. 
Among table-based datasets, SEM~\cite{wang-etal-2021-semeval} and TabEvidence~\cite{gupta-etal-2022-right} are most related to our work.
%
%However, SEM contains simple, crowdsourced claims that lack the complexity of scientific texts, while TabEvidence uses two-column Wikipedia tables, making it less representative of the structure found in scientific literature.
However, SEM features simplified, crowd-generated claims, while TabEvidence is limited to two-column Wikipedia tables, lacking the complexity of scientific tables.
%
%Some frameworks, such as Chain-of-Table~\cite{wang2024chainoftable} and Dater~\cite{10.1145/3539618.3591708}, also incorporate evidence selection in their approaches.
%
%However, they are only evaluated on the claim label prediction task, without assessing the quality of the selected evidence—making the results less trustworthy.
%
Recent frameworks like Chain-of-Table~\cite{wang2024chainoftable} and Dater~\cite{10.1145/3539618.3591708} include evidence selection steps, but only report label accuracy, without evaluating the relevance or quality of the selected evidence, limiting trust in their predictions.

In contrast, our work emphasizes explanation via alignment, explicitly evaluating whether the model selects the correct table cells needed for verification, providing a more faithful and interpretable assessment of reasoning.

\section{Dataset Creation}

In this section, we first briefly introduce the existing SciTab dataset, on which our work is based.
We then describe the process of obtaining the extended version, SciTabAlign, with cell-level explanations.
Finally, we propose a taxonomy of five common ambiguity types, which we hope future work considers to build more reliable claim verification datasets.
We note that we remove all ambiguous cases from our dataset, reducing the number of claims from 868 to 372.

\subsection{Base Dataset: SciTab}

We build on SciTab~\cite{lu-etal-2023-scitab}, the only available dataset for claim verification against scientific tables with naturally occurring claims.
SciTab is derived from SciGen~\cite{moosavi2021scigen}, a table-to-text generation dataset in which each sample consists of a scientific table and its corresponding textual description.

The dataset contains 1,224 claim–table pairs: 457 supported, 411 refuted, and 356 not enough information (NEI).
Supported claims are sourced from original paper content, while refuted and NEI claims are generated by InstructGPT~\cite{NEURIPS2022_b1efde53} and then manually verified.
The original benchmark defines two settings: binary classification (supported vs. refuted) and three-class classification (supported, refuted, NEI), but focuses only on label prediction, without providing explanations.
%In the next section, we describe how we enhance the SciTab dataset by adding explanation information for each claim.

\subsection{Our Dataset: SciTabAlign}
\label{sec_dataset}

% !htb
\begin{figure}[t]
    \centering
    \includegraphics[width=0.99\linewidth]{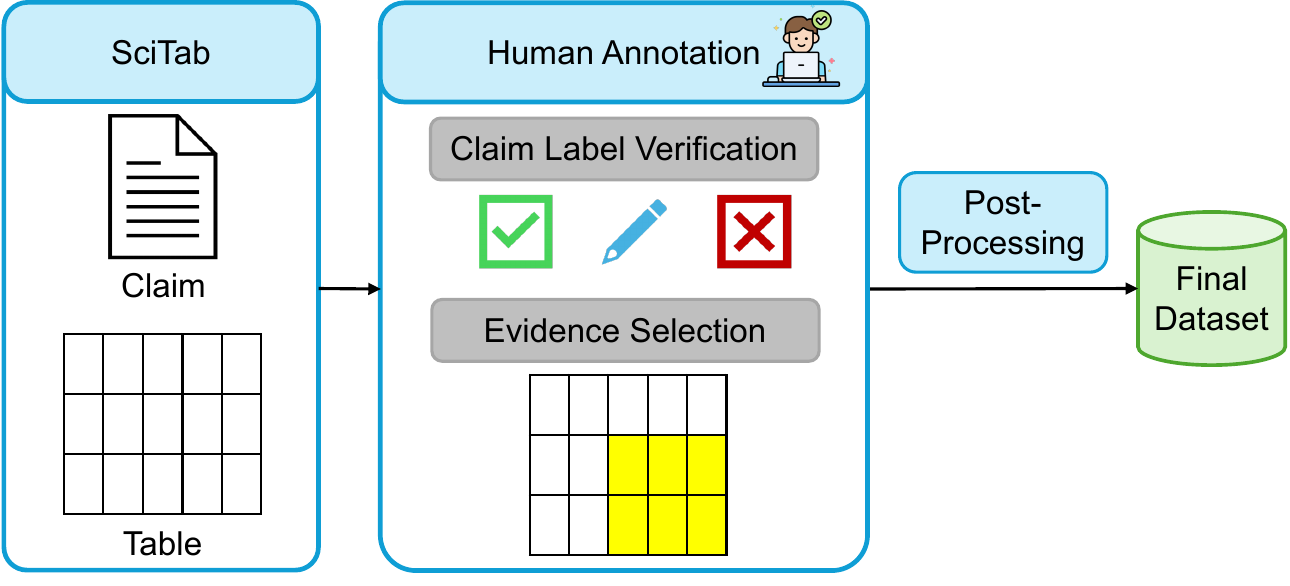}
    \caption{
    Overall dataset creation process.
    }
    \label{fig_overall}
\end{figure}
% SciTabAlign 
We extend SciTab by adding cell-level explanations, i.e.~table regions required to support or refute each claim.
We focus on the supported and refuted claims (868 total), leaving out NEI cases, which are typically under-specified.
As shown in Figure~\ref{fig_overall}, our annotation pipeline includes two tasks: claim label verification and evidence selection.

% \paragraph{Claim Preparation.}
% Given our focus on incorporating explanation information into the claim prediction task, we restrict our selection from the SciTab dataset to claims labeled Supported or Refuted.
% There are 868 claims in total.

\paragraph{Human Annotation.}
Each annotator is given a claim, its label, the associated table, and the caption.
Annotators first verify the correctness of the claim.
If it is clearly supported or refuted, they mark it as \textit{Good}; if the claim is unclear, malformed, or unsupported by the table, they choose \textit{Do Nothing} or optionally revise it (\textit{Revised}).
For \textit{Good} or \textit{Revised} claims, annotators highlight the minimal set of table cells required to determine the label.
Annotation was performed by four NLP researchers (authors of this paper).

\paragraph{Post-Processing.}
After human annotation, we obtain 444 \textit{Good} samples, 81 \textit{Revised}, and 343 \textit{Do Nothing}.
We retain only the samples labeled as \textit{Good}.
Among these, we discard 66 samples in which all table cells are marked (non-informative) and 6 samples containing \textit{NaN} values, resulting in a final dataset of 372 aligned samples (195 supported, 177 refuted).

% We discard samples where all table cells are marked (non-informative) or contain \textit{NaN} values, resulting in a final dataset of 372 aligned samples (195 supported, 177 refuted). 

\paragraph{Inter-Annotator Agreement.}
To assess annotation quality, we conducted a second round of labeling on 50 randomly selected tables (covering 137 claims) by a different annotator.
Using the first annotation as ground truth, we obtained 75.2\% precision, 89.1\% recall, and 78.0\% macro-F1 for cell-level overlap.
%
% These scores indicate strong agreement and compare favorably with prior work on evidence selection~\cite{wang-etal-2021-semeval}.
%
% While not directly comparable, these scores reflect strong agreement and are consistent with prior work such as SEM~\cite{wang-etal-2021-semeval}, which reported 90.1\% exact matches in a similar annotation task.

\subsection{A Proposed Taxonomy}
\label{sec_taxonomy}

%In Section~\ref{sec_dataset}, annotators are given the option to provide a reason when selecting \textit{Do Nothing} for a claim-table pair. 
%Based on these reasons and subsequent discussions among annotators, we proposed a taxonomy of ambiguous cases in the task of verifying scientific claims against tables as follows.

During annotation, we observed frequent edge cases where claim verification was hindered by poor table formatting, unclear language, or missing context.
We propose a taxonomy of five ambiguity types based on annotator notes and discussion:
\begin{itemize}[leftmargin=0pt, itemsep=0cm, topsep=0.1cm]
    \item[] \textbf{(i) Table Conversion Errors:} artifacts introduced during table extraction (e.g.~merged cells, missing entries, formatting loss).
    \item[] \textbf{(ii) Additional Context Requirements:} Claims referencing abbreviations, statistical tests, or assumptions not recoverable from the table alone.
    \item[] \textbf{(iii) Unexpected Claim Types:} Descriptive or meta-level claims (e.g., ``\textit{Table 4 lists the scores of different models.}'') that require no reasoning.
    \item[] \textbf{(iv) Subjective Adjectives:} Use of vague or non-quantifiable terms (e.g., ``\textit{poor}'', ``\textit{substantial}'', or ``\textit{a large margin}'').
    \item[] \textbf{(v) Unclear Claims:} Ambiguous references to table elements (e.g., ``\textit{this model}'', ``\textit{these scores}'').
\end{itemize}

We provide examples for each case in Appendix~\ref{sec_app_dataset}.
We hope this taxonomy will guide future dataset development and improve robustness in scientific claim verification tasks.

%\textbf{(i) Table Conversion Errors:} Errors introduced during the conversion of tables from PDF or LaTeX to text format. 
%These may include incorrect formatting due to merged cells, missing values (NaN), or empty cells.
%\textbf{(ii) Additional Context Requirements:} Cases where the table and its caption alone are insufficient to verify the claim. For example, the table may use unexplained abbreviations or refer to statistical tests without providing necessary details.
%\textbf{(iii) Unexpected Claim Types:} Some claims, such as descriptive statements (e.g., ``Table 4 lists the EM/F1 scores of different models.''), do not require reasoning or verification steps like comparison or calculation, making them incompatible with typical claim verification tasks.
%\textbf{(iv) Subjective Adjectives:} Ambiguity arising from subjective terms in claims, such as \textit{significantly}, \textit{poor}, \textit{substantial}, or \textit{a large margin}, which are difficult to verify objectively.
%\textbf{(v) Unclear Claims:} Claims containing vague pronouns (e.g., \textit{this}, \textit{it}) or ambiguous references to elements like scores, models, or methods, making interpretation difficult.
%We present examples of these types of ambiguous cases in the Appendix~\ref{sec_app_dataset}.
%We hope our taxonomy will guide future dataset construction efforts by highlighting critical cases that require careful consideration.

\section{Experimental Settings}

\paragraph{Models.}
We use three groups of models in our experiments: Table-based LLMs, Open-source LLMs, and Closed-source LLMs.
For table-based LLMs, we use TAPAS-base and TAPAS-large~\cite{herzig-etal-2020-tapas}, pretrained for reasoning over tabular input.
For open-source LLMs, we use the Instruction-tuned variants of Qwen 2.5~\cite[7B and 72B,][]{qwen2025qwen25technicalreport} and Llama 3.1~\cite[8B and 70B,][]{grattafiori2024llama3herdmodels}.
For closed-source LLMs, we use GPT-4o~\cite{openai2024gpt4ocard}.

\paragraph{Prompting Strategies.}
Our dataset contains two subtasks: (1) claim label prediction and (2) cell-level evidence selection.
We conduct experiments using three prompting strategies: zero-shot, few-shot, and Chain-of-Thought~\cite[CoT;][]{wei2022chainofthought}.
%For few-shot and CoT promptings, we use four samples as demonstrations.
%These samples are selected from the revised sample subset, which is not included in our final evaluation set.
For few-shot and CoT promptings, we use four demonstration examples selected from the revised subset of samples not included in the evaluation set, ensuring fair evaluation.

\paragraph{Tabular Representation.}
%\citet{wang2024chainoftable} showed that PIPE encoding with tags (e.g., Col and Row 1) performs better than HTML, TSV, and Markdown formats. 
%Therefore, we use the PIPE encoding method in our experiments.
Following~\citet{wang2024chainoftable}, who found that the PIPE encoding format with explicit tags (e.g.~Col and Row 1) outperformed HTML, TSV, and Markdown formats for tabular data, we adopt PIPE encoding for all of our experiments.

\paragraph{Evaluation Metrics.}
We use Macro-F1 to evaluate both tasks in our dataset: (1) claim label prediction and (2) cell-level evidence selection.
For the claim label prediction task, we compare the predicted label with the ground-truth label.
Our dataset contains two labels: \textit{Supported} and \textit{Refuted}.
For the cell-level evidence selection task, both ground-truth and predicted evidence are represented as lists of (row, column) index tuples using PIPE encoding.
For example, (1, 2) refers to the cell in row 1, column 2.
We compare the two lists to obtain True Positives, False Positives, and False Negatives, and then calculate precision, recall, and F1 based on these values.
The Macro-F1 score is computed as the average of the individual F1 scores.

\section{Results}
All results are shown in Table~\ref{tab_main_result}.
It is noted that, due to cost constraints, we only run GPT-4o on 100 samples selected to match the label distribution of the entire dataset.

\begin{table}[htb!]
    \centering
    \resizebox{\columnwidth}{!}{%
    \begin{tabular}{l | rrr | rrr}
    \toprule
    \multirow{2}{*}{\textbf{Model}} &
    
    \multicolumn{3}{c}{\textbf{Claim Labeling}} &
    \multicolumn{3}{c}{\textbf{Cell Selection}}  \\ 

    \cmidrule(rl){2-4}
    \cmidrule(rl){5-7}

    ~ & Zero & Few & CoT & Zero & Few & CoT \\
    \midrule

TAPAS-base & 48.1 & - & - & - & - & - \\ 
TAPAS-large & 51.6 & - & - & - & - & - \\

\midrule
Llama 3.1 8B & 53.2 & 59.5 & 62.4 & 23.6 & 22.3 & 22.6 \\ 
Llama 3.1 70B & 75.2 & 75.0 & 73.9 & 31.8 & 28.8 & 36.8 \\ 

Qwen 2.5 7B & 66.3 & 68.1 & 67.9 & 20.7 & 16.6 & 17.0 \\ 
Qwen 2.5 72B & 83.5 & 84.7 & 81.5 & 32.8 & 46.7 & \textbf{50.8} \\ 

\midrule
GPT-4o & \textbf{88.4} & 87.0 & 88.0 & 32.4 & 32.9 & 34.8 \\

\bottomrule

\end{tabular}
    }
    \caption{
    Macro-F1 scores of the models on our dataset. 
    `Zero', `Few', and `CoT' denote zero-shot, few-shot, and CoT prompting, respectively.
    }
    \label{tab_main_result}
\end{table}

\paragraph{Claim Prediction Results.}
As expected, GPT-4o achieves the highest score. Larger models, such as Qwen 2.5 72B and Llama 3.1 70B, outperform their smaller 7B and 8B counterparts, and all LLMs surpass the performance of the previous table-based model, TAPAS. 
We also observe that few-shot and CoT prompting are less effective for larger, well-instructed models like the 70B variants and GPT-4o on this familiar label classification task, but remain beneficial for smaller models.

\paragraph{Evidence Selection Results.}

Compared to claim label prediction, evidence cell selection is a more challenging task that most LLMs are unfamiliar with. 
The input consists of a claim and a table, and the output is a list of cell positions—each defined by a row and column index—required to determine the claim’s label. 
This structured output format adds complexity, and overall, all models struggle to achieve high scores on this task.
% Some comments about the scores 
In the zero-shot setting, GPT-4o, Llama 3.1 70B, and Qwen 2.5 72B achieve comparable scores. 
Under few-shot and CoT prompting, GPT-4o's performance remains relatively stable, while Qwen 2.5 72B sees an 18.0 F1 improvement from zero-shot to CoT. 
CoT prompting also boosts Llama 3.1 70B's performance. 
In contrast, smaller models (7B–8B) show decreased performance under both few-shot and CoT prompting compared to zero-shot.

Overall, compared to the human agreement score (78.0 macro F1), the best model still falls short, indicating room for improvement on this task. 
Despite its difficulty and the possibility of multiple valid reasoning paths, our proposed evidence cells can be seen as a minimal, useful set for claim verification. 
In the era of black-box LLMs, focusing solely on the final label is insufficient—evidence selection is equally important for explainability. 
Our work takes a first step toward more interpretable evaluation and highlights the underlying reasoning abilities of models.

\begin{table}[htb!]
    \centering
    \resizebox{\columnwidth}{!}{%
   
\begin{tabular}{l r r r }
\toprule
\textbf{Model}  &  \textbf{Table} & \textbf{Exp.} & \textbf{Table + Exp.}  \\ \midrule
    Llama 3.1 8B & 53.2 & 56.9 & 63.0 \\ 
    
    Llama 3.1 70B & 75.2 & 80.1 & 80.9 \\ 
    
    Qwen 2.5 7B & 66.3 & 67.5 & 69.8 \\ 
    
    Qwen 2.5 72B & 83.5 & 80.6 & 81.9 \\ 
\toprule
\end{tabular}

    }
    \caption{
   Macro-F1 scores of the models on our dataset using different types of input table contexts.
   ``Exp.'' refers to our explanation table cells.
   For all experiments in this table, we use zero-shot prompting.
    }
    \label{tab_useful}
\end{table}

\paragraph{Divergent Effects of Few-Shot and CoT Prompting on Claim Labeling vs. Cell Selection.}
The claim labeling task is a binary classification problem (supported vs. refuted), which closely aligns with tasks that most LLMs are already exposed to during pretraining. 
In contrast, cell selection is a novel task with a different structure, likely unfamiliar to most models.
We observe that for claim labeling, few-shot and CoT prompting benefit smaller models (7B–8B), while larger models (70B–72B) show little to no improvement, likely due to their stronger inherent reasoning capabilities. 
For cell selection, however, smaller models struggle with few-shot and CoT prompting, possibly because the demonstrations are not easily generalizable for this unfamiliar task. 
Larger models perform better in this setting, suggesting greater adaptability to task structure even when it deviates from pretraining distributions.

\paragraph{Effectiveness of Our Explanation Cells.}
To evaluate the effectiveness of our explanation cells, we assess models under two different settings: (1) using only our explanation table cells, and (2) using both the original table and our explanation table cells.
The results are shown in Table~\ref{tab_useful}. 
On average, we observe that using only our explanation cells or combining them with the original table leads to improved task performance.

\section{Analyses}
% \paragraph{Trustworthy Analyses.}

To better understand the correlation between the claim label prediction task and the cell evidence selection task, we categorize outcomes into four types: Correct–Correct, Correct–Incorrect, Incorrect–Correct, and Incorrect–Incorrect.
For claim label prediction, correctness is easily determined based on whether the predicted label (Supported or Refuted) matches the ground truth.
In contrast, cell evidence selection involves list-based predictions, making exact matches more challenging. Therefore, we consider two evaluation criteria: exact match (EM) and a relaxed case where an F1 score of 50.0 or higher is considered correct.

\begin{table}[htb!]
    \centering
    \resizebox{\columnwidth}{!}{%
    \begin{tabular}{l l r r r r r}
\toprule
\textbf{Claim} & \textbf{Cell} & \textbf{L 8B} & \textbf{L 70B} & \textbf{Q 7B} & \textbf{Q 72B} & \textbf{GPT} \\

\midrule

& & \multicolumn{5}{c}{\textbf{Setting 1: Exact Match for Both Tasks}} \\

\cmidrule(rl){3-7}
C & C & 0.0 & 0.0 & 0.0 & 4.6 & 0.0 \\ 
C & I & 63.4 & 73.9 & 68.0 & 73.4 & 88.0 \\ 

I & C & 0.0 & 0.0 & 0.0 & 0.0 & 0.0 \\ 
I & I & 36.6 & 26.1 & 32.0 & 22.0 & 12.0 \\ 
\midrule

& & \multicolumn{5}{c}{\textbf{Setting 2: F1 >= 50 in Cell Selection}} \\
\cmidrule(rl){3-7}

C & C & 10.5 & 26.1 & 4.3 & 44.1 & 30.0 \\

C & I & 53.0 & 47.8 & 63.7 & 33.9 & 58.0 \\ 

I & C & 5.6 & 10.5 & 2.7 & 8.9 & 7.0 \\ 

I & I & 30.9 & 15.6 & 29.3 & 13.2 & 5.0 \\ 

\bottomrule
\end{tabular}
    }
    \caption{
    Categorical statistics (\%) showing the correlation between the claim label prediction and cell evidence selection tasks.
    \textit{C} and \textit{I} denote Correct and Incorrect, respectively.
    \textit{L} and \textit{Q} denote Llama and Qwen, respectively.
    %
    % In the first setting, we use EM for both tasks to determine correctness.
    % In the second setting, we apply EM to the claim label prediction task and consider a cell selection correct if its F1 score is 50.0 or higher.
    The results are from CoT prompting.
    }
    \label{tab_analyses}
\end{table}

The percentage distribution of these cases is shown in Table~\ref{tab_analyses}.
The case where both tasks are correct (C–C) is what we expect.
However, as shown in the table, none of the models achieve a percentage of 50\% for this case—even in the second setting, where an F1 score of 50.0 or higher is considered correct for the cell selection task.
This suggests that while models often predict the claim label correctly, they lack the ability to select the minimal subset of table cells necessary to support that prediction.

\section{Conclusion}
In this work, we highlighted the limitations of scientific claim verification systems that focus solely on label prediction, arguing for the importance of interpretability through evidence selection. 
By reframing table-text alignment as an explanation task and introducing a new dataset with human-annotated cell-level rationales, we provide a more rigorous benchmark for evaluating model reasoning. 
Additionally, we proposed a taxonomy of ambiguous cases in claim verification against tables, which can support future work on dataset construction.
Our findings demonstrate that while LLMs often predict the correct claim labels, they frequently fail to identify the minimal supporting evidence, revealing a gap between accuracy and faithful reasoning. 
This underscores the need for future work to prioritize not just correctness, but also alignment with human-understandable rationales in scientific fact verification tasks.

\section*{Limitations}
Our work has several limitations. 
First, the annotation scale is modest, with 868 claims as input and only 372 claims in the final dataset, which may affect the statistical reliability and generalizability of the findings.
Second, the dataset originates from a specific domain (computer science), which may limit its applicability to tables and claims from other domains.
Third, the PIPE encoding method used may not be well-suited for handling complex table structures, suggesting the need for more robust encoding approaches.

\section*{Acknowledgments}
We would like to thank the anonymous reviewers for their feedback and suggestions on the paper.
This work was supported by JSPS KAKENHI Grant Number 24K03231.

\section*{Ethical Statement and Broader Impact}
We built our dataset based on the publicly available SciTab dataset, which is released under the MIT License. 
We respect the terms of this license and provide appropriate attribution to the original authors. 
To extend the dataset, four NLP researchers manually annotated the data. We created and followed a detailed annotation guideline to ensure consistency, clarity, and fairness in the annotation process. The dataset does not include any personal or sensitive information.

\section*{Potential Risks}
As this dataset consists of 372 samples for claim verification and is intended primarily for evaluation purposes, the risks are limited. The data does not include personal or sensitive information. However, potential risks include biases in the sample selection, which may affect the representativeness of the dataset and the generalizability of evaluation results. Additionally, the dataset could be misapplied outside its intended scope, leading to misleading conclusions if used as a training resource rather than for evaluation.

% We aim to support scientific claim verification against tables with this work. By enriching an existing dataset with high-quality, expert-generated annotations, we enable the development of more effective models for scientific knowledge extraction and reasoning.

\bibliography{anthology,custom}

\appendix

\section{Dataset Creation}
\label{sec_app_dataset}

\subsection{A Proposed Taxonomy}
We present examples for our proposed taxonomy in Section~\ref{sec_taxonomy} in Tables~\ref{tab_ex_i}, \ref{tab_ex_ii}, \ref{tab_ex_iii}, \ref{tab_ex_iv}, and \ref{tab_ex_v}, respectively.

\begin{table*}[ht]
    \centering
    % \resizebox{\textwidth}{!}{%
   \begin{tabular}{@{}p{3cm} p{12cm}@{}}
\toprule

\textbf{Claim} & Comparing POS and SEM tagging (Table 5), we note that higher layer representations do not necessarily improve SEM tagging, while POS tagging does not peak at layer 1. We noticed no improvements in both translation (+0.9 BLEU) and POS and SEM tagging (up to +0.6\% accuracy) when using features extracted from an NMT model trained with residual connections (Table 5). \\

\textbf{Label} & Refuted \\

\textbf{Table Caption} & Table 5: POS and SEM tagging accuracy with features from different layers of 4-layer Uni/Bidirectional/Residual NMT encoders, averaged over all non-English target languages. \\

\textbf{Table} &
% {@{}l@{}}
\begin{tabular}[t]{@{} l c c @{}} 
\texttt{Uni | POS | 0 87.9 | 1 92.0 | 2 91.7 | 3 91.8 | 4 91.9} \\
\texttt{Uni | SEM | 81.8   | 87.8   | 87.4   | 87.6   | 88.2  } \\
\texttt{Bi  | POS | 87.9   | 93.3   | 92.9   | 93.2   | 92.8  } \\
\texttt{Bi  | SEM | 81.9   | 91.3   | 90.8   | 91.9   | 91.9  } \\
\texttt{Res | POS | 87.9   | 92.5   | 91.9   | 92.0   | 92.4  } \\
\texttt{Res | SEM | 81.9   | 88.2   | 87.5   | 87.6   | 88.5  } 

\end{tabular} \\

\bottomrule
\end{tabular}
    % }
    \caption{
    Example of (i) Table Conversion Errors.
    The column headers are merged with the data values. For example, "\texttt{0 87.9}" incorrectly combines the column name \texttt{0} and the value \texttt{87.9}.
    }
    \label{tab_ex_i}
\end{table*}

\begin{table*}[ht]
    \centering
    % \resizebox{\textwidth}{!}{%
   \begin{tabular}{@{}p{3cm} p{12cm}@{}}
\toprule

\textbf{Claim} & After removing the graph attention module, our model gives 24.9 BLEU points. \\

\textbf{Label} & Supported \\

\textbf{Table Caption} & Table 9: Ablation study for modules used in the graph encoder and the LSTM decoder \\

\textbf{Table} &
% {@{}l@{}}
\begin{tabular}[t]{@{} l c c @{}} 
\texttt{[BOLD] Model                    | B       | C      } \\
\texttt{DCGCN4                          | 25.5    | 55.4   } \\
\texttt{Encoder Modules                 | [EMPTY] | [EMPTY]} \\
\texttt{-Linear Combination             | 23.7    | 53.2   } \\
\texttt{-Global Node                    | 24.2    | 54.6   } \\
\texttt{-Direction Aggregation          | 24.6    | 54.6   } \\
\texttt{-Graph Attention                | 24.9    | 54.7   } \\
\texttt{-Global Node \&Linear Combination | 22.9    | 52.4   } \\
\texttt{Decoder Modules                 | [EMPTY] | [EMPTY]} \\
\texttt{-Coverage Mechanism             | 23.8    | 53.0   } 
\end{tabular} \\

\bottomrule
\end{tabular}
    % }
    \caption{
    Example of (ii) Additional Context Requirements.
    \texttt{B} and \texttt{C} stand for BLEU and CHRF++, respectively, but this cannot be inferred from the claim, caption, or table alone. It requires additional context from the original paper.
    }
    \label{tab_ex_ii}
\end{table*}

\begin{table*}[ht]
    \centering
    % \resizebox{\textwidth}{!}{%
    \begin{tabular}{@{}p{3cm} p{12cm}@{}}
\toprule

\textbf{Claim} & Table 4 lists the EM/F1 score of different models. \\

\textbf{Label} & Supported \\

\textbf{Table Caption} & Table 4: Exact match/F1-score on SQuad dataset. “\#Params”: the parameter number of Base. rnet*: results published by Wang et al. (2017). \\

\textbf{Table} &
% {@{}l@{}}
\begin{tabular}[t]{@{} l c c @{}} 
\texttt{Model | \#Params | Base                | +Elmo                     } \\
\texttt{rnet* | -       | 71.1/79.5           | -/-                       } \\
\texttt{LSTM  | 2.67M   | [BOLD] 70.46/78.98  | 75.17/82.79               } \\
\texttt{GRU   | 2.31M   | 70.41/ [BOLD] 79.15 | 75.81/83.12               } \\
\texttt{ATR   | 1.59M   | 69.73/78.70         | 75.06/82.76               } \\
\texttt{SRU   | 2.44M   | 69.27/78.41         | 74.56/82.50               } \\
\texttt{LRN   | 2.14M   | 70.11/78.83         | [BOLD] 76.14/ [BOLD] 83.83} 

\end{tabular} \\

\bottomrule
\end{tabular}
    % }
    \caption{
    Example of (iii) Unexpected Claim Types.
    The claim simply describes what the table shows, similar
to the caption, and does not require any reasoning or data to support it.
    }
    \label{tab_ex_iii}
\end{table*}

\begin{table*}[ht]
    \centering
    \resizebox{\textwidth}{!}{%
   \begin{tabular}{@{}p{1cm} p{15cm}@{}}
\toprule

\textbf{Claim} & [CONTINUE] RELIS significantly outperforms the other RL-based systems. \\

\textbf{Label} & Supported \\

\textbf{Caption} & Table 3: Results of non-RL (top), cross-input (DeepTD) and input-specific (REAPER) RL approaches (middle) compared with RELIS.\\

\textbf{Table} &
% {@{}l@{}}
\small
\begin{tabular}[t]{@{} l c c @{}} 
\texttt{[EMPTY]  | DUC’01 R1 | DUC’01 R2 | DUC’02 R1 | DUC’02 R2 | DUC’04 R1 | DUC’04 R2} \\
\texttt{ICSI     | 33.31                      | 7.33                       | 35.04                      | 8.51                       | 37.31                      | 9.36                      } \\
\texttt{PriorSum | 35.98                      | 7.89                       | 36.63                      | 8.97                       | 38.91                      | 10.07                     } \\
\texttt{TCSum    | <bold>36.45</bold>         | 7.66                       | 36.90                      | 8.61                       | 38.27                      | 9.66                      } \\
\texttt{TCSum-   | 33.45                      | 6.07                       | 34.02                      | 7.39                       | 35.66                      | 8.66                      } \\
\texttt{SRSum    | 36.04                      | 8.44                       | <bold>38.93</bold>         | <bold>10.29</bold>         | 39.29                      | 10.70                     } \\
\texttt{DeepTD   | 28.74                      | 5.95                       | 31.63                      | 7.09                       | 33.57                      | 7.96                      } \\
\texttt{REAPER   | 32.43                      | 6.84                       | 35.03                      | 8.11                       | 37.22                      | 8.64                      } \\
\texttt{RELIS    | 34.73                      | <bold>8.66</bold>          | 37.11                      | 9.12                       | <bold>39.34</bold>         | <bold>10.73</bold>        } 
\end{tabular} \\

\bottomrule
\end{tabular}
    }
    \caption{
    Example of (iv) Subjective Adjectives.
    Whether the performance is considered ``significant'' depends on how the term is defined. Moreover, many argue that using the word ``significant'' requires the result to pass some form of statistical test.
    The original version of the first row is: \texttt{[EMPTY]  | DUC’01 <italic>R</italic>1 | DUC’01 <italic>R</italic>2 | DUC’02 <italic>R</italic>1 | DUC’02 <italic>R</italic>2 | DUC’04 <italic>R</italic>1 | DUC’04 <italic>R</italic>2}.
    }
    \label{tab_ex_iv}
\end{table*}

\begin{table*}[ht]
    \centering
    \resizebox{\textwidth}{!}{%
   \begin{tabular}{@{}p{1.4cm} p{15cm}@{}}
\toprule

\textbf{Claim} & It closely matches the performance of ORACLE with only 0.40\% absolute difference. \\

\textbf{Label} & Supported \\

\textbf{Caption} & Table 3: Accuracy of transferring between aspects. Models with † use labeled data from source aspects. Models with ‡ use human rationales on the target aspect. \\

\textbf{Table} &
% {@{}l@{}}
\small
\begin{tabular}[t]{@{} l c c @{}} 
\texttt{Source            | Target      | Svm   | Ra-Svm‡ | Ra-Cnn‡ | Trans† | Ra-Trans‡† | Ours‡†       | Oracle†} \\
\texttt{Beer aroma+palate | Beer look   | 74.41 | 74.83   | 74.94   | 72.75  | 76.41      | [BOLD] 79.53 | 80.29  } \\
\texttt{Beer look+palate  | Beer aroma  | 68.57 | 69.23   | 67.55   | 69.92  | 76.45      | [BOLD] 77.94 | 78.11  } \\
\texttt{Beer look+aroma   | Beer palate | 63.88 | 67.82   | 65.72   | 74.66  | 73.4       | [BOLD] 75.24 | 75.5   } 

\end{tabular} \\

\bottomrule
\end{tabular}
    }
    \caption{
    Example of (v) Unclear Claims. 
    It is unclear what entity the pronoun ``it'' refers to.
    }
    \label{tab_ex_v}
\end{table*}

\end{document}